\pdfoutput=1
\documentclass[sigconf,balance=false]{acmart}
\usepackage{amsmath}
\usepackage{colortbl}
\usepackage{float}
\usepackage{makecell, array}
\usepackage{tabularx}
\usepackage{xspace}

\newcommand{\fref}[1]{Fig.~\ref{#1}}
\definecolor{FirstPlace}{RGB}{214, 239, 221}
\definecolor{SecondPlace}{RGB}{222, 235, 250}
\definecolor{ThirdPlace}{RGB}{255, 241, 204}
\newcommand{\firstplace}[1]{\cellcolor{FirstPlace}#1}
\newcommand{\secondplace}[1]{\cellcolor{SecondPlace}#1}
\newcommand{\thirdplace}[1]{\cellcolor{ThirdPlace}#1}

\newcommand{\methodname}{G3AR\xspace}

\newcommand{\fullname}{Graph-Guided Neural Visual Geometry for Aerial Registration\xspace}

\newcommand{\papertitle}{\methodname: Graph-Guided Neural Visual Geometry for Scalable Multi-Sequence Aerial Registration}
\newcommand{\suppcite}[1]{\cite{#1}}
\newcommand{\bibliographystylesupp}[1]{}
\newcommand{\bibliographysupp}[1]{}
\AtBeginDocument{%
  \providecommand\BibTeX{{%
    \normalfont B\kern-0.5em{\scshape i\kern-0.25em b}\kern-0.8em\TeX}}}

\copyrightyear{2026}
\acmYear{2026}
\setcopyright{cc}
\setcctype{by}
\acmConference[SA Technical Communications '26]{SIGGRAPH Asia 2026 Technical Communications}{December 01--04, 2026}{Kuala Lumpur, Malaysia}
\acmBooktitle{SIGGRAPH Asia 2026 Technical Communications (SA Technical Communications '26), December 01--04, 2026, Kuala Lumpur, Malaysia}
\acmDOI{10.1145/3829339.3847824}
\acmISBN{979-8-4007-2841-9/2026/12}

\begin{document}
\raggedbottom
\title[G3AR]{\papertitle}

\author{Jeng Wen Joshua Lean}
\orcid{0009-0009-8814-2620}
\affiliation{%
  \institution{National Tsing Hua University}
  \city{Hsinchu}
  \country{Taiwan}}
\email{joshualeanjw@gmail.com}

\author{Ting-Yu Yen}
\orcid{0000-0001-9901-6608}
\affiliation{%
  \institution{National Tsing Hua University}
  \city{Hsinchu}
  \country{Taiwan}}
\email{tingyus995@gmail.com}

\author{Wei-Fang Sun}
\orcid{0000-0003-1276-6946}
\affiliation{%
  \institution{NVIDIA AI Technology Center}
  \country{Taiwan}}
\email{johnsons@nvidia.com}

\author{Simon See}
\orcid{0000-0002-4958-9237}
\affiliation{%
  \institution{NVIDIA AI Technology Center}
  \country{Singapore}}
\email{ssee@nvidia.com}

\author{Hung-Kuo Chu}
\orcid{0000-0001-7153-4411}
\affiliation{%
  \institution{National Tsing Hua University}
  \city{Hsinchu}
  \country{Taiwan}}
\email{hkchu@cs.nthu.edu.tw}

\author{Shih-Hsuan Hung}
\authornote{Corresponding author.}
\orcid{0000-0003-4487-8483}
\affiliation{%
  \institution{National Tsing Hua University}
  \city{Hsinchu}
  \country{Taiwan}}
\email{hungsh@cs.nthu.edu.tw}

\renewcommand{\shortauthors}{Lean et al.}

\begin{abstract}

Full-context neural visual geometry is impractical for thousands of images,
while sequence-based chunking poorly captures irregular non-local overlap in
multi-sequence aerial collections. We present \fullname (\methodname), a
graph-guided framework for scalable dense neural geometry. Before local
inference, \methodname builds a geometrically verified image-proximity graph
that guides bounded overlapping chunks and induces a chunk graph whose maximum
spanning tree defines alignment topology. Compatible backbones process chunks
independently; shared-image predictions then estimate three-dimensional
similarity (Sim(3)) transforms that register local cameras and geometry in a
common frame. Across four real aerial scenes, \methodname improves pose error
and runtime in matched VGGT- and Pi3-backed comparisons, while its DA3 variant
achieves the lowest pose error among evaluated neural-geometry methods.

\end{abstract}



\settopmatter{printacmref=true}
\settopmatter{printfolios=false}


\begin{teaserfigure}
\includegraphics[width=0.9\textwidth]{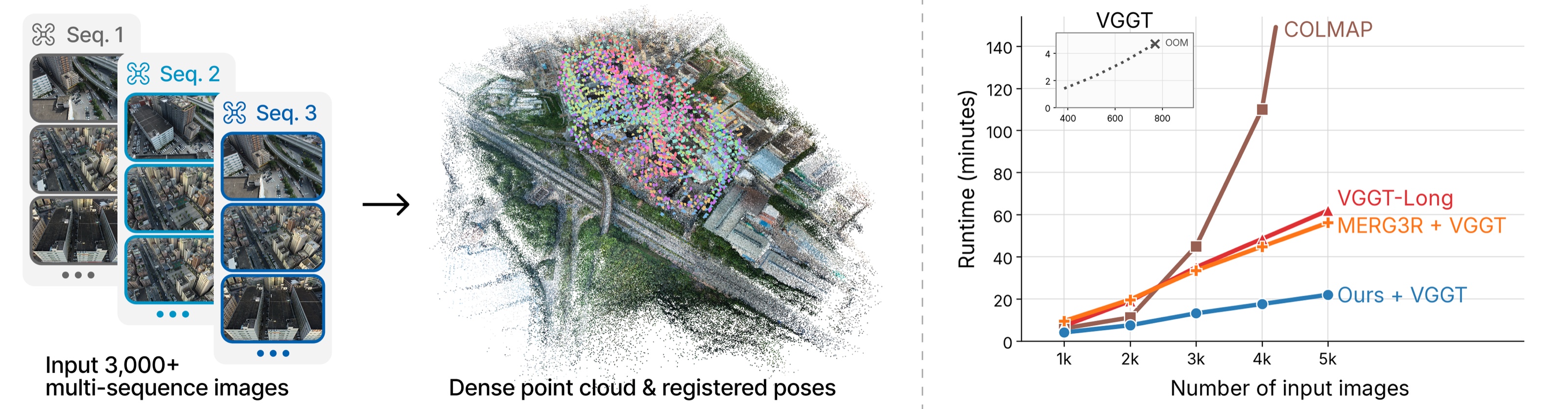}
\centering
\Description{More than 3,000 aerial images from multiple flight sequences are registered into a single dense point cloud. A runtime plot shows Ours plus VGGT scaling more efficiently than VGGT-Long plus VGGT, MERG3R plus VGGT, full-context VGGT, and COLMAP as the collection grows.}
\caption{\textbf{\methodname} registers thousands of multi-sequence aerial images into a coherent point cloud with lower runtime than prior long-context pipelines.}
  \label{fig:teaser}
\end{teaserfigure}

\maketitle

\section{Introduction}

Feed-forward neural geometry models such as VGGT~\shortcite{wang2025vggt},
Pi3~\shortcite{wang2025pi3}, and DA3~\shortcite{lin2025depthanything3}
directly predict camera poses and dense geometry. Yet full-context inference over
thousands of images is impractical. Long-context methods such as
VGGT-Long~\shortcite{deng2025vggtlong} and MERG3R~\shortcite{cheng2026merg3r}
therefore process and align chunks, making results depend on chunk composition
and alignment topology.

Large aerial surveys span multiple flights without a reliable global temporal
order: overlap crosses strips, revisits, or vehicles, while sequence boundaries
may be unrelated. Scaling therefore requires reliable non-local connections for
chunk formation and alignment without exhaustive matching or full-scene
optimization.

\begin{figure*}[!t]
\centering
\includegraphics[width=\textwidth]{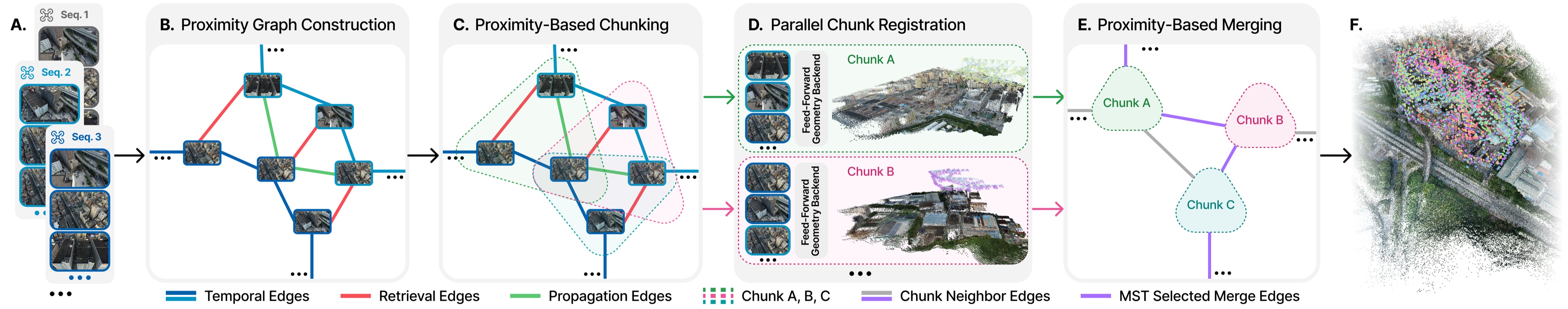}
\Description{A six-panel overview of G3AR. Multiple aerial image sequences are organized into a verified proximity graph, partitioned into overlapping chunks, and processed independently by a feed-forward geometry backbone. A maximum spanning tree of the proximity-derived chunk graph selects pairwise alignments, which are composed to produce registered cameras and aligned dense geometry.}
\caption{Overview of \textbf{\methodname}. \textbf{(A)} Multiple aerial image sequences are provided as input. \textbf{(B)} Verified temporal, retrieval, and propagation edges organize the images into a proximity graph. \textbf{(C)} Graph partitioning and boundary expansion form bounded overlapping chunks. \textbf{(D)} A feed-forward geometry backbone predicts chunk-local camera poses and dense geometry in parallel. \textbf{(E)} A maximum spanning tree of the proximity-derived chunk graph selects shared-image Sim(3) alignments. \textbf{(F)} Composing the transformations registers the cameras and aligns the geometry in a common frame.}
\label{fig:method-overview}
\end{figure*}

We present \textbf{G3AR}: \textbf{G}raph-\textbf{G}uided Neural Visual
\textbf{G}eometry for Scalable Multi-Sequence \textbf{A}erial
\textbf{R}egistration, which adapts graph partitioning to dense neural geometry.
A geometrically verified proximity graph guides bounded overlapping chunks
for neural inference; their chunk graph defines a maximum spanning
tree for shared-image Sim(3) alignment.
Compatible backbones independently predict local cameras and geometry,
supporting multi-GPU inference.
Assembly requires no backbone retraining, global pose-graph refinement,
or post-assembly bundle adjustment.
On four real aerial scenes, \methodname reduces pose error over matched
VGGT- and Pi3-backed baselines, while DA3 achieves the lowest evaluated
neural pose error.
On the 5,621-image synthetic Small City scene from
MatrixCity~\shortcite{li2023matrixcity}, its VGGT variant reduces pose error,
point-cloud error, and runtime relative to VGGT-Long and MERG3R.
With DA3, eight L40 GPUs achieve a $2.29\times$ mean end-to-end speedup
across the four real scenes.

\section{Related Work}

COLMAP~\shortcite{schonberger2016sfm} and GLOMAP~\shortcite{pan2024glomap} combine verified matching, camera registration, and global optimization.
GraphSfM~\shortcite{chen2020graph} partitions verified view graphs into overlapping clusters and merges independent reconstructions.
FastMap~\shortcite{li2025fastmap} and InstantSfM~\shortcite{zhong2025instantsfm} target efficiency; the supplement compares these systems and DAGSfM with \methodname.
Feed-forward models VGGT~\shortcite{wang2025vggt}, Pi3~\shortcite{wang2025pi3}, and DA3~\shortcite{lin2025depthanything3} predict cameras and dense geometry, while VGGT-Long~\shortcite{deng2025vggtlong}, SwiftVGGT~\shortcite{lee2025swiftvggt}, and MERG3R~\shortcite{cheng2026merg3r} scale through subset processing and alignment.
\methodname adapts graph partitioning to neural geometry, using verified aerial proximity to jointly guide bounded inference chunks and inter-chunk Sim(3) alignment.
\section{Method}

Given images $\{I_i\}_{i=1}^{N}$ from one or more flight sequences, \methodname builds a weighted proximity graph $\mathcal{G}=(\mathcal{V},\mathcal{E},w)$, where $\mathcal{V}=[N]$ indexes images, $\mathcal{E}$ contains geometrically verified temporal, retrieval, and propagation edges, and $w_{ij}$ measures verified overlap.
As illustrated in \fref{fig:method-overview},
\textbf{(A)} multi-sequence aerial images are provided;
\textbf{(B)} verified edges form a proximity graph;
\textbf{(C)} each connected component is partitioned and expanded into bounded overlapping chunks $\{\mathcal{C}_k\}$;
\textbf{(D)} a compatible feed-forward geometry backbone processes chunks independently in parallel;
\textbf{(E)} a maximum spanning tree of the induced chunk graph selects shared-image Sim(3) alignments; and
\textbf{(F)} composing these transformations registers cameras and aligns geometry.

\subsection{Proximity Graph Construction}

We propose \emph{temporal}, \emph{retrieval}, and \emph{propagation} edges to recover local and non-local overlap without exhaustive pairing; geometric verification filters repetitive-structure matches and determines edge retention and weight.
For image $i$, flight metadata and UltraVPR~\cite{chen2025ultravpr} provide neighborhoods $\mathcal N_{\mathrm{temp}}(i)$ and $\mathcal N_{\mathrm{ret}}(i)$, respectively, yielding initial candidates
$\widetilde{\mathcal E}_0=\{\{i,j\}:j\in\mathcal N_{\mathrm{temp}}(i)\cup\mathcal N_{\mathrm{ret}}(i)\}$.
Temporal candidates preserve flight continuity; retrieval identifies revisits and cross-strip or cross-sequence overlap.
For each candidate $\{i,j\}$, ALIKED~\shortcite{zhao2023aliked} and LightGlue~\shortcite{lindenberger2023lightglue} produce tentative correspondences $\mathcal M_{ij}$. Fundamental-matrix verification with USAC~\shortcite{raguram2013usac} and MAGSAC~\shortcite{barath2020magsac} yields inliers $\widehat{\mathcal M}_{ij}$.
Let $\rho_i(\widehat{\mathcal M}_{ij})$ denote the fraction of grid cells in $I_i$ containing an inlier. We define
$w_{ij}=|\widehat{\mathcal M}_{ij}|\max\{\rho_i(\widehat{\mathcal M}_{ij}),\rho_j(\widehat{\mathcal M}_{ij})\}$,
with $w_{ij}=0$ upon verification failure.
The operation $\operatorname{Verify}(\cdot)$ retains pairs with $w_{ij}\geq\tau$, where $\tau=200$, giving
$\mathcal E_0=\operatorname{Verify}(\widetilde{\mathcal E}_0)$ and
$\mathcal G_0=(\mathcal V,\mathcal E_0,w)$.
For each retained retrieval proposal $i\!\rightarrow\!j$, propagation proposes edges from $i$ to unconnected neighbors of $j$ in $\mathcal G_0$. Applying the same procedure gives
$\mathcal E=\mathcal E_0\cup\operatorname{Verify}(\widetilde{\mathcal E}_{\mathrm{prop}})$.
UltraVPR only retrieves candidates; verified local geometry supports every final edge. We process $\mathcal G=(\mathcal V,\mathcal E,w)$ independently per connected component.

\begin{table*}[!t]
\caption{Average camera pose estimation on four real aerial scenes. Values are Sim(3)-aligned and averaged equally across scenes. COLMAP and GLOMAP are unranked classical references. Among neural-geometry methods, the top-3 results are highlighted as \protect\colorbox{FirstPlace}{\strut first}, \protect\colorbox{SecondPlace}{\strut second}, and \protect\colorbox{ThirdPlace}{\strut third}.}
\label{tab:real-results}
\centering
\resizebox{0.7\textwidth}{!}{%
\begin{tabular}{lrrrrr}
\hline
Method & ATE RMSE $\downarrow$ & ATE mean $\downarrow$ & ATE median $\downarrow$ & Runtime $\downarrow$ & Max VRAM (GiB) \\
\hline
COLMAP~\cite{schonberger2016sfm} & 0.031 & 0.022 & 0.016 & 61 min 17 s & N/A \\
GLOMAP~\cite{pan2024glomap} & 3.570 & 3.207 & 3.177 & 27 min 00 s & N/A \\
\hline
VGGT-Long~\cite{deng2025vggtlong} & 2.355 & 2.084 & 1.994 & 23 min 50 s & 20.5 \\
SwiftVGGT~\cite{lee2025swiftvggt} & 2.370 & 2.073 & 1.870 & 10 min 25 s & 33.5 \\
MERG3R + VGGT~\cite{cheng2026merg3r} & 2.556 & 2.288 & 2.170 & 23 min 54 s & 20.9 \\
MERG3R + Pi3~\cite{cheng2026merg3r} & \thirdplace{1.030} & \thirdplace{0.900} & \thirdplace{0.835} & 21 min 14 s & 19.3 \\
\hline\hline
Ours + VGGT & 1.855 & 1.529 & 1.317 & \secondplace{9 min 42 s} & 21.7 \\
Ours + Pi3 & \secondplace{0.733} & \secondplace{0.483} & \secondplace{0.384} & \firstplace{9 min 38 s} & 19.7 \\
Ours + DA3 & \firstplace{0.653} & \firstplace{0.403} & \firstplace{0.301} & \thirdplace{10 min 23 s} & 26.7 \\
\hline
\end{tabular}%
}
\end{table*}


\subsection{Proximity-Based Chunking}
\label{sec:proximity-chunking}
We form bounded inference chunks that preserve high-proximity neighborhoods and share images for 3D alignment.
Let $b_p$ and $b_c$ denote the partition budget and maximum expanded-chunk size. Weighted METIS partitioning~\shortcite{karypis1998metis} favors low-weight cuts, producing disjoint partitions $\{\mathcal P_k\}$ that cover each component of $\mathcal G$ and satisfy $|\mathcal P_k|\leq b_p$. Components within budget remain unchanged; those without usable internal adjacency use sequential partitioning.
For each $\mathcal P_k$, let $\bar{\mathcal P_k}$ contain outside images connected to it by $\mathcal E$. Adding the highest-proximity images from $\bar{\mathcal P_k}$ forms $\mathcal C_k$, subject to $|\mathcal C_k|\leq b_c$. Shared images $\mathcal O_{k\ell}=\mathcal C_k\cap\mathcal C_\ell$ constrain subsequent Sim(3) alignment.
A compatible feed-forward geometry backbone independently predicts chunk-local camera poses and dense geometry.
Backbone-independent graph construction and chunking support VGGT, Pi3, and DA3, with chunk inference distributable across GPUs.

\subsection{Proximity-Based Merging}
\label{sec:proximity-merging}

After chunk registration, we lift image proximity to a chunk graph for lightweight assembly.
Expanded chunks induce
$\mathcal G_{\mathcal C}=(\mathcal K,\mathcal E_{\mathcal C},\eta)$,
where $\mathcal K=[N_{\mathcal C}]$ indexes chunks and
$\{k,\ell\}\in\mathcal E_{\mathcal C}$ when neighboring chunks share images,
i.e., $\mathcal O_{k\ell}\neq\emptyset$.
Weights $\eta_{k\ell}$ combine proximity evidence and shared-image overlap.
A maximum spanning tree is extracted independently per connected component, prioritizing strong connections without sequential ordering.
We define
$\eta_{k\ell}=|\mathcal O_{k\ell}|\max w_{ij}$
over cross-partition edges.
For each selected edge $\{k,\ell\}$, dense predictions of images in $\mathcal O_{k\ell}$ provide corresponding 3D points in both chunk frames.
Following VGGT-Long~\shortcite{deng2025vggtlong}, \methodname uses iteratively reweighted least squares to estimate a robust pairwise Sim(3) transformation, recovering relative scale, rotation, and translation.
For each tree, we select a reference chunk and compose pairwise Sim(3) transformations along unique tree paths, placing all reachable cameras and geometry in a common frame.
This tree-based assembly requires neither global pose-graph Levenberg-Marquardt refinement nor post-merge bundle adjustment.
\begin{figure*}[!t]
\centering
\includegraphics[width=\textwidth]{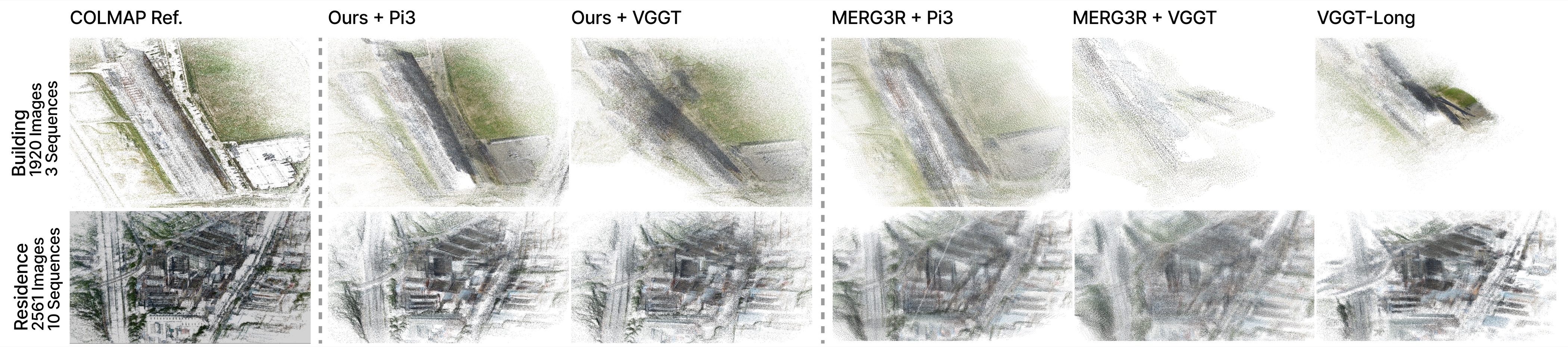}
\Description{Building and Residence point clouds from the COLMAP reference,
Ours plus Pi3, Ours plus VGGT, MERG3R plus Pi3, MERG3R plus VGGT, and VGGT-Long.}
\caption{Building and Residence point clouds under matched local inference
budgets. Graph-guided variants retain broader roof, facade, and ground
extents than the displayed long-context baselines.}
\label{fig:qual-crop}
\end{figure*}

\section{Experiments}
We evaluate Building and Rubble from Mill19~\shortcite{turki2022meganerf}
and Residence and Sci-Art from UrbanScene3D~\shortcite{lin2022urbanscene3d}
(1,657--2,998 images), plus the 5,621-image synthetic Small City scene
from MatrixCity~\shortcite{li2023matrixcity}, with ground-truth poses
and a dense reference point cloud.
We report camera coverage, Sim(3)-aligned ATE, end-to-end runtime,
peak GPU memory, and Chamfer-L1 where available, using one NVIDIA L40
GPU unless noted.
For controlled comparison, VGGT-Long, MERG3R, and \methodname use at most
95 images per chunk with a 35-image overlap target
($b_p=75$, $b_c=95$), retaining their own chunking and alignment strategies.
The supplement provides broader classical SfM comparisons, scaling details,
and ablations of temporal-order dependence, alignment topology, and global optimization.

\noindent
\textbf{Quantitative Results.}
Full-context runs of VGGT~\shortcite{wang2025vggt},
Pi3~\shortcite{wang2025pi3}, DA3~\shortcite{lin2025depthanything3},
FastVGGT~\shortcite{shen2025fastvggt}, and
LiteVGGT~\shortcite{shu2025litevggt} exceed 48~GiB and are omitted.
\methodname reduces ATE and runtime relative to VGGT-Long and MERG3R
with VGGT, and to MERG3R with Pi3.
All three variants maintain full camera coverage; Ours + DA3 achieves
the lowest ATE and Ours + Pi3 the shortest runtime among evaluated neural
methods (Table~\ref{tab:real-results}).
On Small City~\shortcite{li2023matrixcity}, Ours + VGGT achieves the lowest
ATE, Chamfer-L1, and runtime among the compared neural pipelines
(Table~\ref{tab:small-city}).

\begin{table}[!t]
\caption{Small City results with VGGT local geometry. Errors are Sim(3)-aligned in dataset units.}
\label{tab:small-city}
\centering
\resizebox{0.85\columnwidth}{!}{%
\begin{tabular}{lrrr}
\hline
Method & ATE RMSE $\downarrow$ & Runtime $\downarrow$
       & Chamfer-L1 $\downarrow$ \\
\hline
VGGT-Long & 6.6277 & 44 min 55 s & 2.7457 \\
MERG3R + VGGT & 5.2523 & 44 min 3 s & 3.1699 \\
Ours + VGGT & \textbf{4.7639} & \textbf{24 min 8 s}
            & \textbf{1.3408} \\
\hline
\end{tabular}
}
\end{table}
\begin{table}[!t]
\caption{DA3 chunk-image selection ablation, averaged equally over four real scenes.}
\label{tab:ablation-chunk-selection}
\centering
\begin{tabular}{lr}
\hline
\small{Variant} & \small{ATE RMSE $\downarrow$} \\
\hline
\small{Temporal edges only} & 2.625 \\
\small{Temporal + retrieval edges} & 0.819 \\
\small{Temporal + retrieval + propagation edges} & 0.653 \\
\hline
\end{tabular}
\end{table}

\noindent
\textbf{Qualitative Results.}
On Building and Residence, graph-guided variants preserve coherent global
layouts and broader extents than the displayed long-context baselines,
while local geometry remains more diffuse than COLMAP
(\fref{fig:qual-crop}).

\begin{figure}[!t]
\centering
\includegraphics[width=0.7\columnwidth]{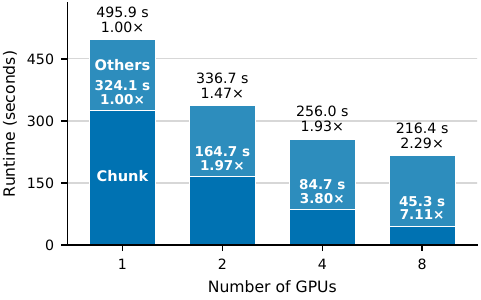}
\Description{Mean runtime across four real scenes for one, two, four,
and eight GPUs. Chunk-registration time falls nearly inversely with GPU
count; graph, merge, and evaluation time remain nearly constant.}
\caption{Scaling over four real scenes. Labels show mean runtime and
mean per-scene speedup relative to one GPU.}
\label{fig:multi-gpu-scaling}
\end{figure}

\noindent
\textbf{Multi-GPU Scaling.}
Across four real scenes with DA3, eight warmed L40 GPUs achieve
$7.11\times$ mean chunk speedup at 88.9\% efficiency and
$2.29\times$ mean end-to-end speedup (\fref{fig:multi-gpu-scaling}).

\noindent
\textbf{Ablations.}
Chunk-selection ablations with DA3 across four real scenes show that
retrieval reduces ATE RMSE over temporal edges alone, with further gains
from propagation (Table~\ref{tab:ablation-chunk-selection}),
supporting verified non-local edges for chunk selection.

\section{Conclusion}
\methodname uses graph-guided partitioning, shared-image overlap, and inter-chunk
Sim(3) alignment to scale dense neural geometry for multi-sequence aerial
registration. Across four real scenes, all three backbones retain full coverage
and improve pose error over matched long-context counterparts; eight-GPU chunk
inference achieves 7.11$\times$ mean speedup (2.29$\times$ end-to-end). Future
work will improve graph robustness beyond aerial scenes and support incremental
reconstruction.

\begin{acks}
We thank the anonymous reviewers. This work was supported by Taiwan's National
Science and Technology Council grants 114-2221-E-007-114-MY3,
113-2221-E-007-102-MY3, 114-2221-E-007-115-MY3, and 115-2634-F-007-003.
We also thank NVIDIA Corporation and NVIDIA AI Technology Center (NVAITC) for
providing access to the Taipei-1 supercomputer.
\end{acks}

\bibliographystyle{ACM-Reference-Format}
\bibliography{ref}


\begin{thebibliography}{22}


\ifx \showCODEN    \undefined \def \showCODEN     #1{\unskip}     \fi
\ifx \showISBNx    \undefined \def \showISBNx     #1{\unskip}     \fi
\ifx \showISBNxiii \undefined \def \showISBNxiii  #1{\unskip}     \fi
\ifx \showISSN     \undefined \def \showISSN      #1{\unskip}     \fi
\ifx \showLCCN     \undefined \def \showLCCN      #1{\unskip}     \fi
\ifx \shownote     \undefined \def \shownote      #1{#1}          \fi
\ifx \showarticletitle \undefined \def \showarticletitle #1{#1}   \fi
\ifx \showURL      \undefined \def \showURL       {\relax}        \fi
\providecommand\bibfield[2]{#2}
\providecommand\bibinfo[2]{#2}
\providecommand\natexlab[1]{#1}
\providecommand\showeprint[2][]{arXiv:#2}

\bibitem[Barath et~al\mbox{.}(2020)]%
        {barath2020magsac}
\bibfield{author}{\bibinfo{person}{Daniel Barath}, \bibinfo{person}{Jana
  Noskova}, \bibinfo{person}{Maksym Ivashechkin}, {and}
  \bibinfo{person}{Ji{\v{r}}{\'i} Matas}.} \bibinfo{year}{2020}\natexlab{}.
\newblock \showarticletitle{{MAGSAC}++, a Fast, Reliable and Accurate Robust
  Estimator}. In \bibinfo{booktitle}{\emph{Proceedings of the IEEE/CVF
  Conference on Computer Vision and Pattern Recognition}}.
  \bibinfo{pages}{1304--1312}.
\newblock


\bibitem[Chen et~al\mbox{.}(2025)]%
        {chen2025ultravpr}
\bibfield{author}{\bibinfo{person}{Chao Chen}, \bibinfo{person}{Chunyu Li},
  \bibinfo{person}{Mengfan He}, \bibinfo{person}{Jun Wang},
  \bibinfo{person}{Fei Xing}, {and} \bibinfo{person}{Ziyang Meng}.}
  \bibinfo{year}{2025}\natexlab{}.
\newblock \showarticletitle{{UltraVPR}: Unsupervised Lightweight
  Rotation-Invariant Aerial Visual Place Recognition}.
\newblock \bibinfo{journal}{\emph{IEEE Robotics and Automation Letters}}
  \bibinfo{volume}{10}, \bibinfo{number}{9} (\bibinfo{year}{2025}),
  \bibinfo{pages}{9096--9103}.
\newblock
\href{https://doi.org/10.1109/LRA.2025.3592075}{doi:\nolinkurl{10.1109/LRA.2025.3592075}}


\bibitem[Chen et~al\mbox{.}(2020)]%
        {chen2020graph}
\bibfield{author}{\bibinfo{person}{Yu Chen}, \bibinfo{person}{Shuhan Shen},
  \bibinfo{person}{Yisong Chen}, {and} \bibinfo{person}{Guoping Wang}.}
  \bibinfo{year}{2020}\natexlab{}.
\newblock \showarticletitle{Graph-based Parallel Large Scale Structure from
  Motion}.
\newblock \bibinfo{journal}{\emph{Pattern Recognition}}  \bibinfo{volume}{107}
  (\bibinfo{year}{2020}), \bibinfo{pages}{107537}.
\newblock
\href{https://doi.org/10.1016/j.patcog.2020.107537}{doi:\nolinkurl{10.1016/j.patcog.2020.107537}}


\bibitem[Cheng et~al\mbox{.}(2026)]%
        {cheng2026merg3r}
\bibfield{author}{\bibinfo{person}{Leo~Kaixuan Cheng}, \bibinfo{person}{Abdus
  Shaikh}, \bibinfo{person}{Ruofan Liang}, \bibinfo{person}{Zhijie Wu},
  \bibinfo{person}{Yushi Guan}, {and} \bibinfo{person}{Nandita Vijaykumar}.}
  \bibinfo{year}{2026}\natexlab{}.
\newblock \showarticletitle{{MERG3R}: A Divide-and-Conquer Approach to
  Large-Scale Neural Visual Geometry}. In \bibinfo{booktitle}{\emph{Proceedings
  of the IEEE/CVF Conference on Computer Vision and Pattern Recognition}}.
  \bibinfo{pages}{28969--28978}.
\newblock


\bibitem[Deng et~al\mbox{.}(2026)]%
        {deng2025vggtlong}
\bibfield{author}{\bibinfo{person}{Kai Deng}, \bibinfo{person}{Zexin Ti},
  \bibinfo{person}{Jiawei Xu}, \bibinfo{person}{Jian Yang}, {and}
  \bibinfo{person}{Jin Xie}.} \bibinfo{year}{2026}\natexlab{}.
\newblock \showarticletitle{{VGGT-Long}: Chunk it, Loop it, Align it -- Pushing
  {VGGT}'s Limits on Kilometer-scale Long {RGB} Sequences}. In
  \bibinfo{booktitle}{\emph{Proceedings of the IEEE International Conference on
  Robotics and Automation}}.
\newblock


\bibitem[Karypis and Kumar(1998)]%
        {karypis1998metis}
\bibfield{author}{\bibinfo{person}{George Karypis} {and} \bibinfo{person}{Vipin
  Kumar}.} \bibinfo{year}{1998}\natexlab{}.
\newblock \showarticletitle{A Fast and High Quality Multilevel Scheme for
  Partitioning Irregular Graphs}.
\newblock \bibinfo{journal}{\emph{SIAM Journal on Scientific Computing}}
  \bibinfo{volume}{20}, \bibinfo{number}{1} (\bibinfo{year}{1998}),
  \bibinfo{pages}{359--392}.
\newblock
\href{https://doi.org/10.1137/S1064827595287997}{doi:\nolinkurl{10.1137/S1064827595287997}}


\bibitem[Lee et~al\mbox{.}(2026)]%
        {lee2025swiftvggt}
\bibfield{author}{\bibinfo{person}{Jungho Lee}, \bibinfo{person}{Minhyeok Lee},
  \bibinfo{person}{Sunghun Yang}, \bibinfo{person}{Minseok Kang}, {and}
  \bibinfo{person}{Sangyoun Lee}.} \bibinfo{year}{2026}\natexlab{}.
\newblock \showarticletitle{{SwiftVGGT}: A Scalable Visual Geometry Grounded
  Transformer for Large-Scale Scenes}. In \bibinfo{booktitle}{\emph{Proceedings
  of the IEEE/CVF Conference on Computer Vision and Pattern Recognition
  Findings}}. \bibinfo{pages}{447--456}.
\newblock


\bibitem[Li et~al\mbox{.}(2026)]%
        {li2025fastmap}
\bibfield{author}{\bibinfo{person}{Jiahao Li}, \bibinfo{person}{Haochen Wang},
  \bibinfo{person}{Muhammad~Zubair Irshad}, \bibinfo{person}{Igor Vasiljevic},
  \bibinfo{person}{Matthew~R. Walter}, \bibinfo{person}{Vitor~Campagnolo
  Guizilini}, {and} \bibinfo{person}{Greg Shakhnarovich}.}
  \bibinfo{year}{2026}\natexlab{}.
\newblock \showarticletitle{{FastMap}: Revisiting Structure from Motion through
  First-Order Optimization}. In \bibinfo{booktitle}{\emph{Proceedings of the
  International Conference on 3D Vision}}. \bibinfo{pages}{29--39}.
\newblock
\href{https://doi.org/10.1109/3DV69130.2026.00010}{doi:\nolinkurl{10.1109/3DV69130.2026.00010}}


\bibitem[Li et~al\mbox{.}(2023)]%
        {li2023matrixcity}
\bibfield{author}{\bibinfo{person}{Yixuan Li}, \bibinfo{person}{Lihan Jiang},
  \bibinfo{person}{Linning Xu}, \bibinfo{person}{Yuanbo Xiangli},
  \bibinfo{person}{Zhenzhi Wang}, \bibinfo{person}{Dahua Lin}, {and}
  \bibinfo{person}{Bo Dai}.} \bibinfo{year}{2023}\natexlab{}.
\newblock \showarticletitle{{MatrixCity}: A Large-scale City Dataset for
  City-scale Neural Rendering and Beyond}. In
  \bibinfo{booktitle}{\emph{Proceedings of the IEEE/CVF International
  Conference on Computer Vision}}. \bibinfo{pages}{3205--3215}.
\newblock


\bibitem[Lin et~al\mbox{.}(2025)]%
        {lin2025depthanything3}
\bibfield{author}{\bibinfo{person}{Haotong Lin}, \bibinfo{person}{Sili Chen},
  \bibinfo{person}{Jun~Hao Liew}, \bibinfo{person}{Donny~Y. Chen},
  \bibinfo{person}{Zhenyu Li}, \bibinfo{person}{Guang Shi},
  \bibinfo{person}{Jiashi Feng}, {and} \bibinfo{person}{Bingyi Kang}.}
  \bibinfo{year}{2025}\natexlab{}.
\newblock \bibinfo{title}{Depth Anything 3: Recovering the Visual Space from
  Any Views}.
\newblock \bibinfo{howpublished}{arXiv preprint arXiv:2511.10647}.
\newblock


\bibitem[Lin et~al\mbox{.}(2022)]%
        {lin2022urbanscene3d}
\bibfield{author}{\bibinfo{person}{Liqiang Lin}, \bibinfo{person}{Yilin Liu},
  \bibinfo{person}{Yue Hu}, \bibinfo{person}{Xingguang Yan},
  \bibinfo{person}{Ke Xie}, {and} \bibinfo{person}{Hui Huang}.}
  \bibinfo{year}{2022}\natexlab{}.
\newblock \showarticletitle{Capturing, Reconstructing, and Simulating: The
  {UrbanScene3D} Dataset}. In \bibinfo{booktitle}{\emph{Proceedings of the
  European Conference on Computer Vision}}. \bibinfo{pages}{93--109}.
\newblock
\href{https://doi.org/10.1007/978-3-031-20074-8_6}{doi:\nolinkurl{10.1007/978-3-031-20074-8_6}}


\bibitem[Lindenberger et~al\mbox{.}(2023)]%
        {lindenberger2023lightglue}
\bibfield{author}{\bibinfo{person}{Philipp Lindenberger},
  \bibinfo{person}{Paul-Edouard Sarlin}, {and} \bibinfo{person}{Marc
  Pollefeys}.} \bibinfo{year}{2023}\natexlab{}.
\newblock \showarticletitle{{LightGlue}: Local Feature Matching at Light
  Speed}. In \bibinfo{booktitle}{\emph{Proceedings of the IEEE/CVF
  International Conference on Computer Vision}}. \bibinfo{pages}{17627--17638}.
\newblock


\bibitem[Pan et~al\mbox{.}(2024)]%
        {pan2024glomap}
\bibfield{author}{\bibinfo{person}{Linfei Pan}, \bibinfo{person}{Daniel
  Barath}, \bibinfo{person}{Marc Pollefeys}, {and} \bibinfo{person}{Johannes~L.
  Sch{\"o}nberger}.} \bibinfo{year}{2024}\natexlab{}.
\newblock \showarticletitle{Global Structure-from-Motion Revisited}. In
  \bibinfo{booktitle}{\emph{Proceedings of the European Conference on Computer
  Vision}}. \bibinfo{pages}{58--77}.
\newblock
\href{https://doi.org/10.1007/978-3-031-73661-2_4}{doi:\nolinkurl{10.1007/978-3-031-73661-2_4}}


\bibitem[Raguram et~al\mbox{.}(2013)]%
        {raguram2013usac}
\bibfield{author}{\bibinfo{person}{Rahul Raguram},
  \bibinfo{person}{Ond{\v{r}}ej Chum}, \bibinfo{person}{Marc Pollefeys},
  \bibinfo{person}{Ji{\v{r}}{\'i} Matas}, {and} \bibinfo{person}{Jan-Michael
  Frahm}.} \bibinfo{year}{2013}\natexlab{}.
\newblock \showarticletitle{{USAC}: A Universal Framework for Random Sample
  Consensus}.
\newblock \bibinfo{journal}{\emph{IEEE Transactions on Pattern Analysis and
  Machine Intelligence}} \bibinfo{volume}{35}, \bibinfo{number}{8}
  (\bibinfo{year}{2013}), \bibinfo{pages}{2022--2038}.
\newblock
\href{https://doi.org/10.1109/TPAMI.2012.257}{doi:\nolinkurl{10.1109/TPAMI.2012.257}}


\bibitem[Sch{\"o}nberger and Frahm(2016)]%
        {schonberger2016sfm}
\bibfield{author}{\bibinfo{person}{Johannes~L. Sch{\"o}nberger} {and}
  \bibinfo{person}{Jan-Michael Frahm}.} \bibinfo{year}{2016}\natexlab{}.
\newblock \showarticletitle{Structure-from-Motion Revisited}. In
  \bibinfo{booktitle}{\emph{Proceedings of the IEEE Conference on Computer
  Vision and Pattern Recognition}}. \bibinfo{pages}{4104--4113}.
\newblock


\bibitem[Shen et~al\mbox{.}(2026)]%
        {shen2025fastvggt}
\bibfield{author}{\bibinfo{person}{You Shen}, \bibinfo{person}{Zhipeng Zhang},
  \bibinfo{person}{Yansong Qu}, \bibinfo{person}{Xiawu Zheng},
  \bibinfo{person}{Jiayi Ji}, \bibinfo{person}{Shengchuan Zhang}, {and}
  \bibinfo{person}{Liujuan Cao}.} \bibinfo{year}{2026}\natexlab{}.
\newblock \showarticletitle{{FastVGGT}: Fast Visual Geometry Transformer}. In
  \bibinfo{booktitle}{\emph{International Conference on Learning
  Representations}}.
\newblock


\bibitem[Shu et~al\mbox{.}(2026)]%
        {shu2025litevggt}
\bibfield{author}{\bibinfo{person}{Zhijian Shu}, \bibinfo{person}{Cheng Lin},
  \bibinfo{person}{Tao Xie}, \bibinfo{person}{Wei Yin}, \bibinfo{person}{Ben
  Li}, \bibinfo{person}{Zhiyuan Pu}, \bibinfo{person}{Weize Li},
  \bibinfo{person}{Yao Yao}, \bibinfo{person}{Xun Cao},
  \bibinfo{person}{Xiaoyang Guo}, {and} \bibinfo{person}{Xiao-Xiao Long}.}
  \bibinfo{year}{2026}\natexlab{}.
\newblock \showarticletitle{{LiteVGGT}: Boosting Vanilla {VGGT} via
  Geometry-aware Cached Token Merging}. In
  \bibinfo{booktitle}{\emph{Proceedings of the IEEE/CVF Conference on Computer
  Vision and Pattern Recognition}}. \bibinfo{pages}{36422--36432}.
\newblock


\bibitem[Turki et~al\mbox{.}(2022)]%
        {turki2022meganerf}
\bibfield{author}{\bibinfo{person}{Haithem Turki}, \bibinfo{person}{Deva
  Ramanan}, {and} \bibinfo{person}{Mahadev Satyanarayanan}.}
  \bibinfo{year}{2022}\natexlab{}.
\newblock \showarticletitle{{Mega-NeRF}: Scalable Construction of Large-Scale
  {NeRFs} for Virtual Fly-Throughs}. In \bibinfo{booktitle}{\emph{Proceedings
  of the IEEE/CVF Conference on Computer Vision and Pattern Recognition}}.
  \bibinfo{pages}{12922--12931}.
\newblock


\bibitem[Wang et~al\mbox{.}(2025)]%
        {wang2025vggt}
\bibfield{author}{\bibinfo{person}{Jianyuan Wang}, \bibinfo{person}{Minghao
  Chen}, \bibinfo{person}{Nikita Karaev}, \bibinfo{person}{Andrea Vedaldi},
  \bibinfo{person}{Christian Rupprecht}, {and} \bibinfo{person}{David
  Novotny}.} \bibinfo{year}{2025}\natexlab{}.
\newblock \showarticletitle{{VGGT}: Visual Geometry Grounded Transformer}. In
  \bibinfo{booktitle}{\emph{Proceedings of the IEEE/CVF Conference on Computer
  Vision and Pattern Recognition}}. \bibinfo{pages}{5294--5306}.
\newblock


\bibitem[Wang et~al\mbox{.}(2026)]%
        {wang2025pi3}
\bibfield{author}{\bibinfo{person}{Yifan Wang}, \bibinfo{person}{Jianjun Zhou},
  \bibinfo{person}{Haoyi Zhu}, \bibinfo{person}{Wenzheng Chang},
  \bibinfo{person}{Yang Zhou}, \bibinfo{person}{Zizun Li},
  \bibinfo{person}{Junyi Chen}, \bibinfo{person}{Jiangmiao Pang},
  \bibinfo{person}{Chunhua Shen}, {and} \bibinfo{person}{Tong He}.}
  \bibinfo{year}{2026}\natexlab{}.
\newblock \showarticletitle{$\pi^3$: Permutation-Equivariant Visual Geometry
  Learning}. In \bibinfo{booktitle}{\emph{International Conference on Learning
  Representations}}.
\newblock


\bibitem[Zhao et~al\mbox{.}(2023)]%
        {zhao2023aliked}
\bibfield{author}{\bibinfo{person}{Xiaoming Zhao}, \bibinfo{person}{Xingming
  Wu}, \bibinfo{person}{Weihai Chen}, \bibinfo{person}{Peter C.~Y. Chen},
  \bibinfo{person}{Qingsong Xu}, {and} \bibinfo{person}{Zhengguo Li}.}
  \bibinfo{year}{2023}\natexlab{}.
\newblock \showarticletitle{{ALIKED}: A Lighter Keypoint and Descriptor
  Extraction Network via Deformable Transformation}.
\newblock \bibinfo{journal}{\emph{IEEE Transactions on Instrumentation and
  Measurement}}  \bibinfo{volume}{72} (\bibinfo{year}{2023}),
  \bibinfo{pages}{1--16}.
\newblock
\href{https://doi.org/10.1109/TIM.2023.3271000}{doi:\nolinkurl{10.1109/TIM.2023.3271000}}


\bibitem[Zhong et~al\mbox{.}(2025)]%
        {zhong2025instantsfm}
\bibfield{author}{\bibinfo{person}{Jiankun Zhong}, \bibinfo{person}{Zitong
  Zhan}, \bibinfo{person}{Quankai Gao}, \bibinfo{person}{Ziyu Chen},
  \bibinfo{person}{Haozhe Lou}, \bibinfo{person}{Jiageng Mao},
  \bibinfo{person}{Ulrich Neumann}, \bibinfo{person}{Chen Wang}, {and}
  \bibinfo{person}{Yue Wang}.} \bibinfo{year}{2025}\natexlab{}.
\newblock \bibinfo{title}{{InstantSfM}: Towards {GPU}-Native {SfM} for the Deep
  Learning Era}.
\newblock \bibinfo{howpublished}{arXiv preprint arXiv:2510.13310}.
\newblock


\end{thebibliography}

\clearpage
\onecolumn
\ifdefined\standalonesupplement
\thispagestyle{empty}
\begin{center}
{\Large\bfseries Supplemental Material\par}
\vspace{0.4em}
{\large \papertitle\par}
\medskip
Jeng Wen Joshua Lean\textsuperscript{1}, Ting-Yu Yen\textsuperscript{1}, Wei-Fang Sun\textsuperscript{2}, Simon See\textsuperscript{3},\\
Hung-Kuo Chu\textsuperscript{1}, and Shih-Hsuan Hung\textsuperscript{1}\par
\smallskip
{\small\textsuperscript{1}National Tsing Hua University, Hsinchu, Taiwan\par
\textsuperscript{2}NVIDIA AI Technology Center, Taipei, Taiwan\qquad
\textsuperscript{3}NVIDIA AI Technology Center, Singapore\par
Corresponding author: \href{mailto:hungsh@cs.nthu.edu.tw}{hungsh@cs.nthu.edu.tw}\par}
\vspace{1em}
\end{center}
\fi
\appendix
\ifdefined\standalonesupplement
\renewcommand{\thetable}{S\arabic{table}}
\fi

\section{Classical Structure-from-Motion Comparison}
\label{app:classical-sfm}

We compare \methodname with classical and learning-assisted
structure-from-motion pipelines on the four real aerial scenes.
Coverage is reported because DAGSfM, FastMap, and InstantSfM produce
partial reconstructions; their pose errors are therefore evaluated on
different camera subsets and are not directly comparable with
full-coverage results.

\begin{table}[H]
\caption{Comparison averaged equally across four real aerial scenes.
Coverage is the percentage of input cameras registered. ATE values are
Sim(3)-aligned. For methods marked $^\dagger$, ATE is computed only over
registered cameras when available and is therefore not directly comparable
with full-coverage results. Dashes denote unavailable values.}
\label{tab:classical-sfm}
\centering
\resizebox{0.9\textwidth}{!}{%
\begin{tabular}{lrrrrrr}
\hline
Method
& Coverage
& ATE RMSE $\downarrow$
& ATE mean $\downarrow$
& ATE median $\downarrow$
& Runtime $\downarrow$
& Max VRAM (GiB) \\
\hline
COLMAP~\suppcite{schonberger2016sfm}
& 100.0\% & 0.031 & 0.022 & 0.016 & 61 min 17 s & N/A \\
GLOMAP~\suppcite{pan2024glomap}
& 100.0\% & 3.570 & 3.207 & 3.177 & 27 min 00 s & N/A \\
DAGSfM$^\dagger$~\suppcite{chen2020graph}
& 2.1\% & -- & -- & -- & 104 min 51 s & N/A \\
FastMap$^\dagger$~\suppcite{li2025fastmap}
& 96.4\% & 1.663 & 1.397 & 1.307 & 4 min 26 s & 5.6 \\
InstantSfM$^\dagger$~\suppcite{zhong2025instantsfm}
& 81.1\% & 3.235 & 2.854 & 2.732 & 8 min 55 s & 22.6 \\
\hline\hline
Ours + VGGT
& 100.0\% & 1.855 & 1.529 & 1.317 & 9 min 42 s & 21.7 \\
Ours + Pi3
& 100.0\% & 0.733 & 0.483 & 0.384 & 9 min 38 s & 19.7 \\
Ours + DA3
& 100.0\% & 0.653 & 0.403 & 0.301 & 10 min 23 s & 26.7 \\
\hline
\end{tabular}%
}
\end{table}

Among full-coverage methods, COLMAP has the lowest pose error but the
longest runtime. All \methodname variants finish in approximately ten
minutes; the Pi3- and DA3-backed variants also achieve lower ATE than GLOMAP.

\section{Multi-GPU Strong Scaling}
\label{app:multi-gpu-scaling}

We measure steady-state strong scaling on Building, Rubble, Residence,
and Sci-Art using DA3 and 1, 2, 4, and 8 NVIDIA L40 GPUs. Each point is
the arithmetic mean of one timed run per scene; speedups are computed
per scene against its one-GPU result before equal-weight averaging.
Each scene uses fixed chunks generated with a target/maximum/overlap
configuration of 75/95/35 images. Persistent workers are warmed before
measurement, excluding process launch and model loading. Each proximity
graph is constructed once, and its measured cost is included at every
scale point. Chunks are assigned using image-count-weighted
longest-processing-time scheduling.

\begin{table}[H]
\caption{Strong scaling with DA3, averaged equally across four real
scenes. End-to-end time includes proximity-graph construction, chunk
registration, merging, and evaluation.}
\label{tab:multi-gpu-scaling}
\centering
\resizebox{0.58\textwidth}{!}{%
\begin{tabular}{rrrrrr}
\hline
GPUs
& \makecell{Mean chunk\\time (s)}
& \makecell{Mean end-to-end\\time (s)}
& \makecell{Mean chunk\\speedup}
& \makecell{Mean chunk\\efficiency}
& \makecell{Mean end-to-end\\speedup} \\
\hline
1 & 324.15 & 495.87 & 1.000$\times$ & 100.0\% & 1.000$\times$ \\
2 & 164.71 & 336.68 & 1.967$\times$ & 98.3\% & 1.471$\times$ \\
4 & 84.74 & 256.02 & 3.803$\times$ & 95.1\% & 1.932$\times$ \\
8 & 45.31 & 216.37 & 7.113$\times$ & 88.9\% & 2.289$\times$ \\
\hline
\end{tabular}%
}
\end{table}

At eight GPUs, mean per-scene chunk speedup reaches 7.113$\times$ with
88.9\% efficiency, while mean end-to-end speedup reaches 2.289$\times$.
The approximately 172-second mean graph, merging, and evaluation
remainder therefore becomes the primary scaling bottleneck.

\section{Limitations and Optional Reconnection}
\label{app:graph-connectivity}

All reported experiments use the verification threshold $\tau=200$
without edge promotion. If thresholding disconnects the proximity graph,
its components are reconstructed independently, producing separate
islands whose relative coordinate frames are not determined by the
pipeline. Repetitive structures may also yield incorrect high-weight
edges that survive verification. If selected for tree-based alignment,
their errors can propagate along the tree.

The implementation also supports optional reconnection:
among candidate edges between disconnected components, it promotes the
edge with the highest verification score, even when that score falls
below $\tau$. This option relaxes the acceptance threshold to recover
connectivity, but a promoted edge may be incorrect and introduce
alignment error. It was not used in the reported results.

\section{Additional Ablations}
\label{app:additional-ablations}

\subsection{Temporal-Order Robustness}
\label{app:temporal-order}

We remove temporal candidates on Building to test whether \methodname
requires temporal adjacency (Table~\ref{tab:ablation-temporal-order}).
Standard retrieval registers only 14 of 21 partitions; wider retrieval
restores all 21, with lower ATE but higher runtime than the default.

\begin{table}[H]
\caption{Temporal-order ablation on Building with DA3. Standard and
wide retrieval use $(k,\mathrm{keep},\mathrm{ratio})=(50,5,3)$ and
$(300,40,10)$, respectively. $^\dagger$ATE is computed from a partial
reconstruction and is not directly comparable with full-registration
results.}
\label{tab:ablation-temporal-order}
\centering
\begin{tabularx}{0.65\columnwidth}{Xrrr}
\hline
Variant
& \makecell{Registered\\partitions}
& ATE RMSE $\downarrow$
& Runtime $\downarrow$ \\
\hline
Temp. + Retr. + Prop.
& 21/21 & 0.4192 & 5 min 00 s \\
Retr. + Prop.
& 14/21 & 0.7142$^\dagger$ & 4 min 46 s \\
Wide Retr. + Prop.
& 21/21 & 0.3930 & 7 min 45 s \\
\hline
\end{tabularx}
\end{table}

On this scene, temporal adjacency is an efficient proposal prior rather
than a strict requirement: sufficiently broad retrieval recovers complete
registration, but at higher runtime.

\subsection{Chunk Alignment Topology}
\label{app:alignment-topology}

We compare three alignment topologies on Rubble using identical selected
chunks (Table~\ref{tab:ablation-alignment-topology}). The maximum spanning
tree gives the lowest error, supporting verified chunk proximity as the
criterion for selecting alignment edges in this setting.

\begin{table}[H]
\caption{Chunk alignment-topology ablation on Rubble with DA3 using
identical selected chunks.}
\label{tab:ablation-alignment-topology}
\centering
\begin{tabularx}{0.5\columnwidth}{Xr}
\hline
Variant & ATE RMSE $\downarrow$ \\
\hline
Maximum spanning tree & 0.3227 \\
Temporal chain & 0.5663 \\
Similarity-based Hamiltonian path & 0.4462 \\
\hline
\end{tabularx}
\end{table}

\subsection{Global Optimization}
\label{app:global-optimization}

We compare tree-based assembly with two global-optimization variants
on Rubble (Table~\ref{tab:ablation-global-optimization}). Neither improves
ATE in this ablation, and bundle adjustment substantially increases merge
time, supporting maximum spanning tree assembly as the default.

\begin{table}[H]
\caption{Global-optimization variants on Rubble with DA3.
Levenberg--Marquardt uses all neighboring-chunk alignments as loop
constraints; post-merge bundle adjustment follows the MERG3R settings.}
\label{tab:ablation-global-optimization}
\centering
\begin{tabularx}{0.5\columnwidth}{Xrr}
\hline
Variant
& Merge time $\downarrow$
& ATE RMSE $\downarrow$ \\
\hline
Ours
& 0 min 15 s & 0.3227 \\
Ours + Levenberg--Marquardt
& 0 min 20 s & 0.4432 \\
Ours + bundle adjustment
& 2 min 48 s & 0.4419 \\
\hline
\end{tabularx}
\end{table}

\bibliographystylesupp{ACM-Reference-Format}
\bibliographysupp{ref}

\end{document}